\documentclass[sigconf]{acmart}
\usepackage{svg}
\setsvg{inkscape = inkscape -z -D}
\setsvg{svgpath = images/} 
\usepackage{ctable} % For special rule
\usepackage{amssymb}
\usepackage{amsthm}
\usepackage{amsmath}
\DeclareMathOperator*{\argminA}{\arg\min}
\usepackage[linesnumbered,vlined]{algorithm2e}
\usepackage{subfig}
\usepackage[shortlabels]{enumitem}
\usepackage{textcomp}

\AtBeginDocument{%
  \providecommand\BibTeX{{%
    \normalfont B\kern-0.5em{\scshape i\kern-0.25em b}\kern-0.8em\TeX}}}

 \setcopyright{none}
\renewcommand\footnotetextcopyrightpermission[1]{} 

\newcommand\blfootnote[1]{%
  \begingroup
  \renewcommand\thefootnote{}\footnote{#1}%
  \addtocounter{footnote}{-1}%
  \endgroup
}

\begin{document}

\title{Collective Learning From Diverse Datasets for Entity Typing in the Wild}

\author{Abhishek}
\email{abhishek.abhishek@iitg.ac.in}
\affiliation{%
  \institution{Indian Institute of Technology Guwahati}
  \city{Guwahati}
  \state{Assam}
  \country{India}
}

\author{Amar Prakash Azad}
\email{amarazad@in.ibm.com}
\author{Balaji Ganesan}
\email{bganesa1@in.ibm.com}
\affiliation{%
  \institution{IBM Research Lab}
  \country{India}
  }

\author{Ashish Anand}
\email{anand.ashish@iitg.ac.in}
\author{Amit Awekar}
\email{awekar@iitg.ac.in}
\affiliation{%
  \institution{Indian Institute of Technology Guwahati}
  \city{Guwahati}
  \state{Assam}
  \country{India}
}

\renewcommand{\shortauthors}{Abhishek, et al.}

\begin{abstract}
  Entity typing (ET) is the problem of assigning labels to given entity mentions in a sentence. Existing works for ET require knowledge about the domain and target label set for a given test instance.  ET in the absence of such knowledge is a novel problem that we address as ET in the wild. We hypothesize that the solution to this problem is to build supervised models that generalize better on the ET task as a whole, rather than a specific dataset. In this direction, we propose a Collective Learning Framework (CLF), which enables learning from diverse datasets in a unified way.  The CLF first creates a unified hierarchical label set (UHLS) and a label mapping by aggregating label information from all available datasets. Then it builds a single neural network classifier using UHLS, label mapping and a partial loss function. The single classifier predicts the finest possible label across all available domains even though these labels may not be present in any domain-specific dataset. We also propose a set of evaluation schemes and metrics to evaluate the performance of models in this novel problem.  Extensive experimentation on seven diverse real-world datasets demonstrates the efficacy of our CLF.
\end{abstract}




\maketitle

\section{Introduction}
\blfootnote{Copyright \textcopyright 2019 for this paper by its authors. Use permitted under Creative Commons License Attribution 4.0 International (CC BY 4.0).}

\begin{figure}[t]
\centering
\includesvg[width=1.0\columnwidth]{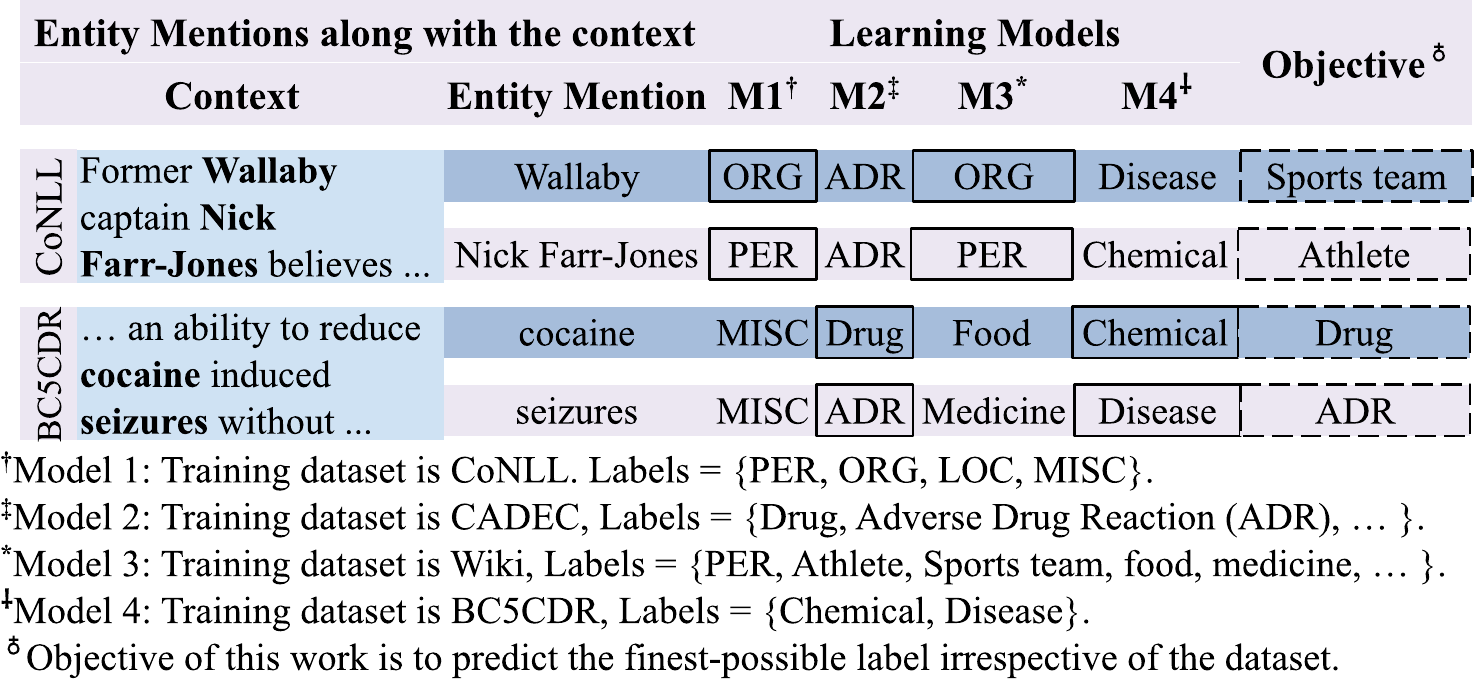}
\caption{The output of four learning models on typing four entity mentions. For example, the model M1 trained on CoNLL dataset assigned the type ORG to the entity mention Wallaby, from the same dataset.}
\label{fig:problem_illustration}
\end{figure}

Evolution of ET has led to the generation of multiple datasets. These datasets differ from each other in terms of their domain or label set or both. Here, the domain of a dataset represents the data distribution of its sentences. The label set represents the entity types annotated.  Existing work for ET requires knowledge of the domain and the target label of a test instance \cite{ren2016afet}. Figure \ref{fig:problem_illustration} illustrates this issue where four learning models are typing four entity mentions. We can observe that, in order to make a reasonable prediction (output with a solid border), it is required to assign labels from a model which has been trained on a dataset with similar domain and labels as that of test instances. However, domain and target label information of a test instance is unknown in several NLP applications such as entity ranking for web question answering systems \cite{dong2015hybrid} and knowledge base completion \cite{45634}, where ET models are used.

We address ET in the absence of domain and target label set knowledge as  ET in the wild problem. As a result, we have to predict the best possible labels for all test instances as illustrated in Figure \ref{fig:problem_illustration} (output with dashed line border). These labels may not be present in the same domain dataset. For example,  the prediction of the label \textit{sports team} for the entity mention Wallaby, when the best possible fine-grained label (\textit{sports team}) is not present in the same domain CoNLL dataset \cite{tjong2003introduction}. We hypothesize that the solution to this problem is to build supervised models that generalize better on the ET task as a whole, rather than a specific dataset. This solution requires collective learning from several diverse datasets. 

\begin{figure}[]
\centering
\includesvg[width=0.85\columnwidth]{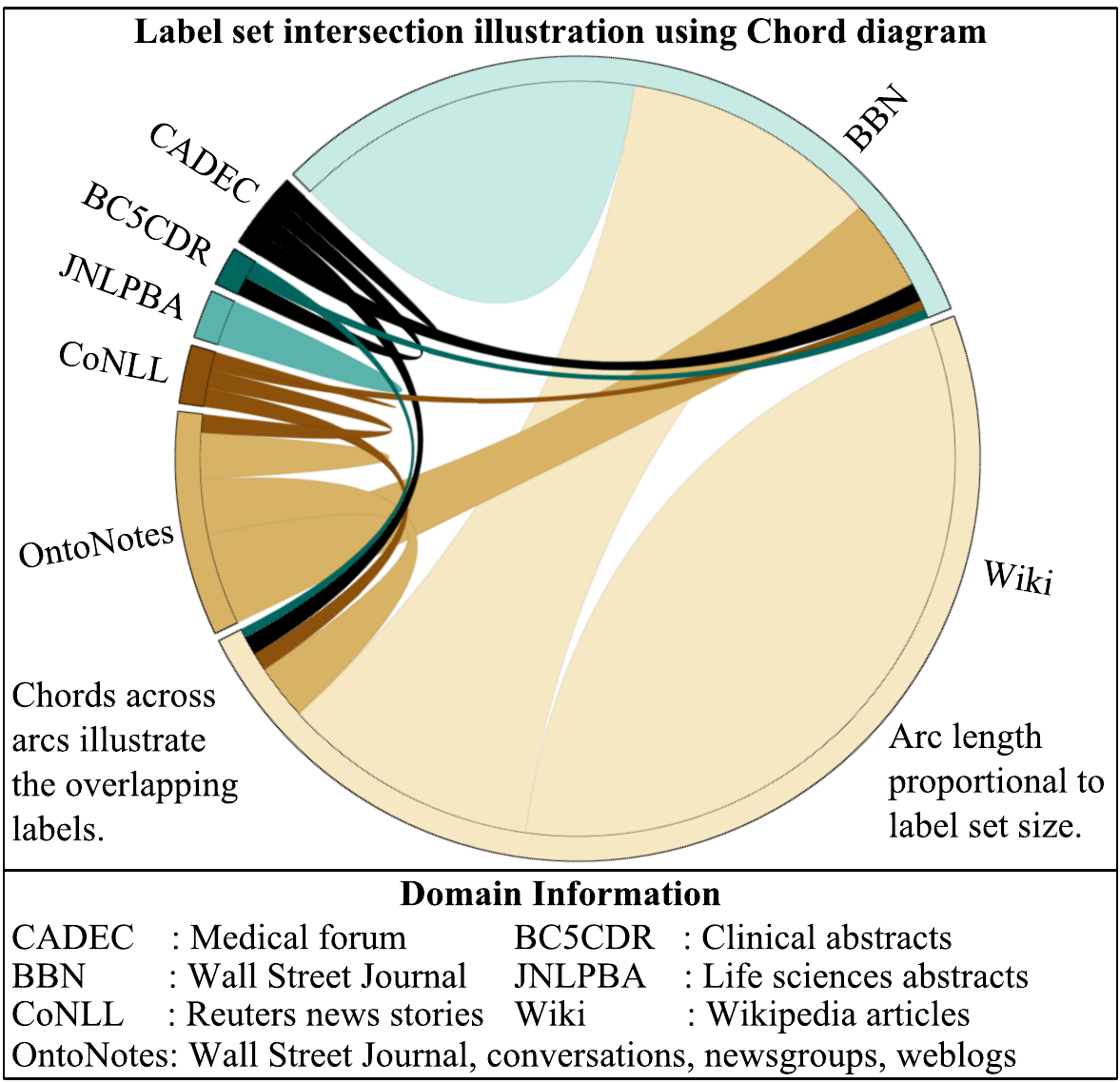}
\caption{Illustration of the diversity of the seven ET datasets in the label set and domain.}
\label{fig:label_overlap}
\end{figure}

However, collectively learning from diverse datasets is a challenging problem. Figure \ref{fig:label_overlap} illustrates the diversity of seven ET datasets. We can observe that every dataset provides some distinct information for the ET task such as domain and labels. For example, CADEC dataset \cite{karimi2015cadec} contains informally written sentences from a medical forum, whereas JNLPBA dataset \cite{kim2004introduction} contains formally written sentences from scientific abstracts in life sciences. Moreover, there is an overlap in the label sets as well as a relation between the labels of these datasets. For example, both CoNLL and Wiki \cite{ling2012fine} datasets have a label \textit{person}. However, only Wiki dataset has a label \textit{athlete}, a subtype of \textit{person}. This means that CoNLL dataset can also contain \textit{athlete} mentions but were only annotated with a coarse label \textit{person}. Thus, learning collectively from these diverse datasets require models to learn a useful feature or representation of the sentences from diverse domains as well as to learn the relation among labels.

This study proposes a collective learning framework for the ET in the wild problem. CLF first builds a unified hierarchical label set (UHLS) and associated label mapping by pooling labels from diverse datasets. Then, a single classifier\footnote{We used the term single classifier to denote a learning model with a single classification head being trained on multiple datasets with different labels together.} collectively learns from the pooled dataset using UHLS, label mapping and a partial hierarchy aware loss function.

In the UHLS, the nodes are contributed by different datasets, and a parent-child relation among nodes translate to a coarse-fine label relation. During construction of UHLS, a mapping from every dataset specific label to the UHLS nodes is also constructed. We expect to have one-to-many mappings, as in the case of real-world datasets. For example, a coarse-grained label for a dataset could be mapped to multiple nodes in the UHLS introduced by some other dataset. During the UHLS construction, human judgment is used when comparing two labels. This effort is four orders of magnitude lesser compared to annotating every dataset with fine-grained labels.

Utilizing the UHLS and the mapping, we can view several domain-specific datasets as a collection of a multi-domain dataset having the same label set. On this combined dataset, we use an LSTM \cite{hochreiter1997long} based encoder to learn a useful representation of the text followed by a partial hierarchical loss function \cite{xu2018neural} for label classification. This setup enables a single neural network classifier to predict fine-grained labels across all domains, even though the fine-grained label was not present in any in-domain dataset. 

We also propose a set of evaluation schemes and metrics for the ET in the wild problem. In our evaluation schemes, we evaluate learning models performance on a test set which is formed by merging test instances of seven diverse datasets. To excel on this merged test set, learning models must generalize beyond a single dataset. Our evaluation metrics are designed to measure learning models performance to predict the best possible fine-grained label. We compared a single classifier model trained with our proposed framework with an ensemble of various models. Our model outperforms competitive baselines with a significant margin.

Our contributions can be highlighted as below:
\begin{enumerate}[topsep=0.3pt]
    \itemsep0em 
    \item We propose a novel problem of ET in the wild with the objective of building better generalizable ET models ($\S$ \ref{sec:problem_desc}).
    \item We propose a novel collective learning framework which makes it possible to train a single classifier on an amalgam of diverse ET datasets, enabling fine-grained prediction across all the datasets, i.e., a generalized model for ET task as a whole ($\S$ \ref{sec:proposed_approach}).
    \item We propose evaluation schemes and evaluation metrics to compare learning models for the ET in the wild problem setting ($\S$ \ref{sec:evaluation_setting}, \ref{sec:evaluation_metric}).
\end{enumerate}

\section{Terminologies and Problem Definition}\label{sec:problem_desc}
In this section, we formally define the ET in the wild problem and related terminologies.

\noindent
\textbf{Dataset:} A dataset, $\mathbb{D}$, is a collection of $(X, \mathcal{D}, \mathcal{Y})$. Here, $X$ corresponds to a corpus of sentences with entity boundaries annotated, $\mathcal{D}$ corresponds to the domain and $\mathcal{Y} = \{y_{1}, \ldots y_{n}\}$ is the set of labels used to annotate each entity mention in the $X$. It is possible that two datasets share domain but differ in their label sets or vice versa. Here the domain means the data characteristics such as writing style and vocabulary. For example, sentences in the CoNLL dataset are sampled from Reuters news stories around 1999, whereas, sentences in the CADEC dataset are from medical forum posts around 2015. Thus, these datasets have different domains.

\begin{figure}[t]
\centering
\includesvg[width=1.0\columnwidth]{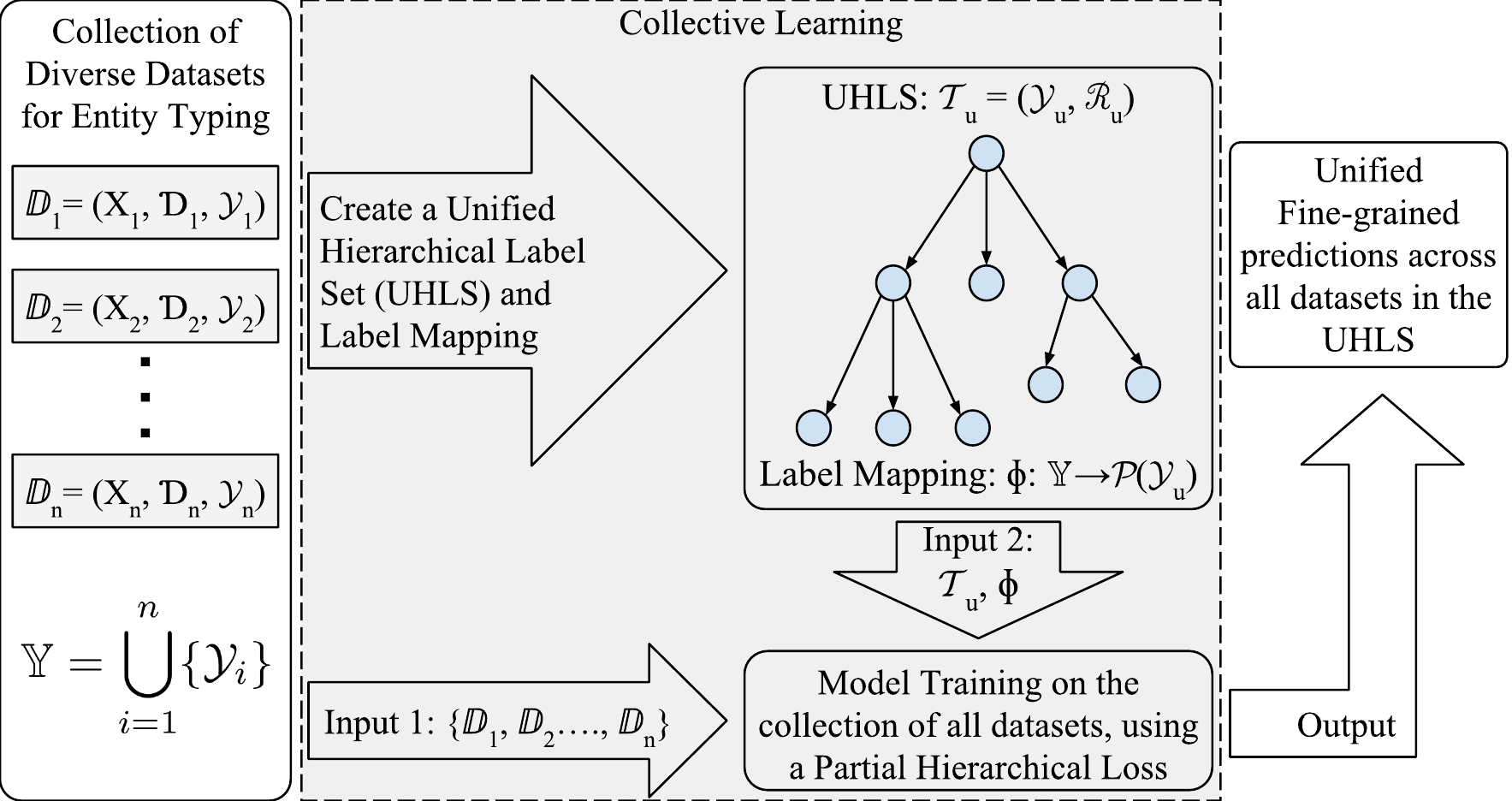}
\caption{An overview of the proposed collective learning framework.}
\label{fig:framework_overview}
\end{figure}

\noindent
\textbf{Label space:} A label space $\mathcal{L}$ for a particular label $y$, is defined as a set of entities that can be assigned a label $y$. For example, the label space for a label \textit{car} includes mentions of all cars including that of label space of differet car types such as \textit{hatchback}, \textit{SUV} etc. For different datasets, even if two labels with the same name exist, their label space can be different. The label space information is defined in the annotation guidelines used to create datasets.

\noindent
\textbf{Type Hierarchy:} A type or label  hierarchy, $\mathcal{T}$, is a natural way to organize label set in a hierarchy. It is formally defined as $(\mathcal{Y}, \mathcal{R})$, where $\mathcal{Y}$ is the type set and $\mathcal{R} = \{(y_{i}, y_{j}) \;|\; y_{i},y_{j} \in \mathcal{Y} \; \& \; i \neq j \; \& \; \mathcal{L}(y_{i}) \prec \mathcal{L}(y_{j}) \}$ is the relation set, in which $(y_{i}, y_{j})$ means that $y_{i}$ is a subtype of $y_{j}$ or in other words the label space of $y_{i}$ is subsumed within the label space of $y_{j}$.

\noindent
\textbf{ET in the Wild problem definition}
Given $n$ datasets, $\mathbb{D}_{1}, \ldots, \mathbb{D}_{n}$, each having its own domain and label set, $\mathcal{D}_{i}$ and $\mathcal{Y}_{i}$ respectively, the objective is to predict the best possible fine-grained label from the set of all labels, $\mathbb{Y} = \bigcup\limits_{i=1}^{n}\{\mathcal{Y}_{i}\}$, for a test entity mention. The fine-grained label might not be present in any in-domain dataset.

\section{Collective Learning Framework (CLF)}\label{sec:proposed_approach}
Figure \ref{fig:framework_overview} provides a complete overview of the CLF, which is based on the following key observations and ideas:
\begin{enumerate}
\itemsep0em 
\item From the set of all available labels $\mathbb{Y}$, it is possible to construct a type hierarchy $\mathcal{T}_{u} = (\mathcal{Y}_{u}, \mathcal{R}_{u})$ where $\mathcal{Y}_{u} \subseteq \mathbb{Y}$ ($\S$ \ref{sec:uhls}). %In $\mathcal{T}_{u}$, the fine-grained labels are present at the leaf level of the hierarchy and non-leaf nodes represents coarse labels. 
\item We can map each $y \in \mathbb{Y}$, to one or more than one node in $\mathcal{T}_{u}$, such that the $\mathcal{L}(y)$ is same as the label space of the union of the mapped nodes ($\S$ \ref{sec:uhls}).
\item Using the above hierarchy and mapping, now even if for some datasets we only have the coarse labels, i.e., the labels which are mapped to non-leaf nodes, a learning model with a partial hierarchy aware loss function can predict fine labels ($\S$ \ref{sec:utp}, \ref{sec:partial_loss}).% by capitalizing on the knowledge learned by simultaneously training on the other datasets whose labels were mapped to the leaf nodes.
\end{enumerate}

\begin{figure}[t]
\centering
\includesvg[width=0.9\columnwidth]{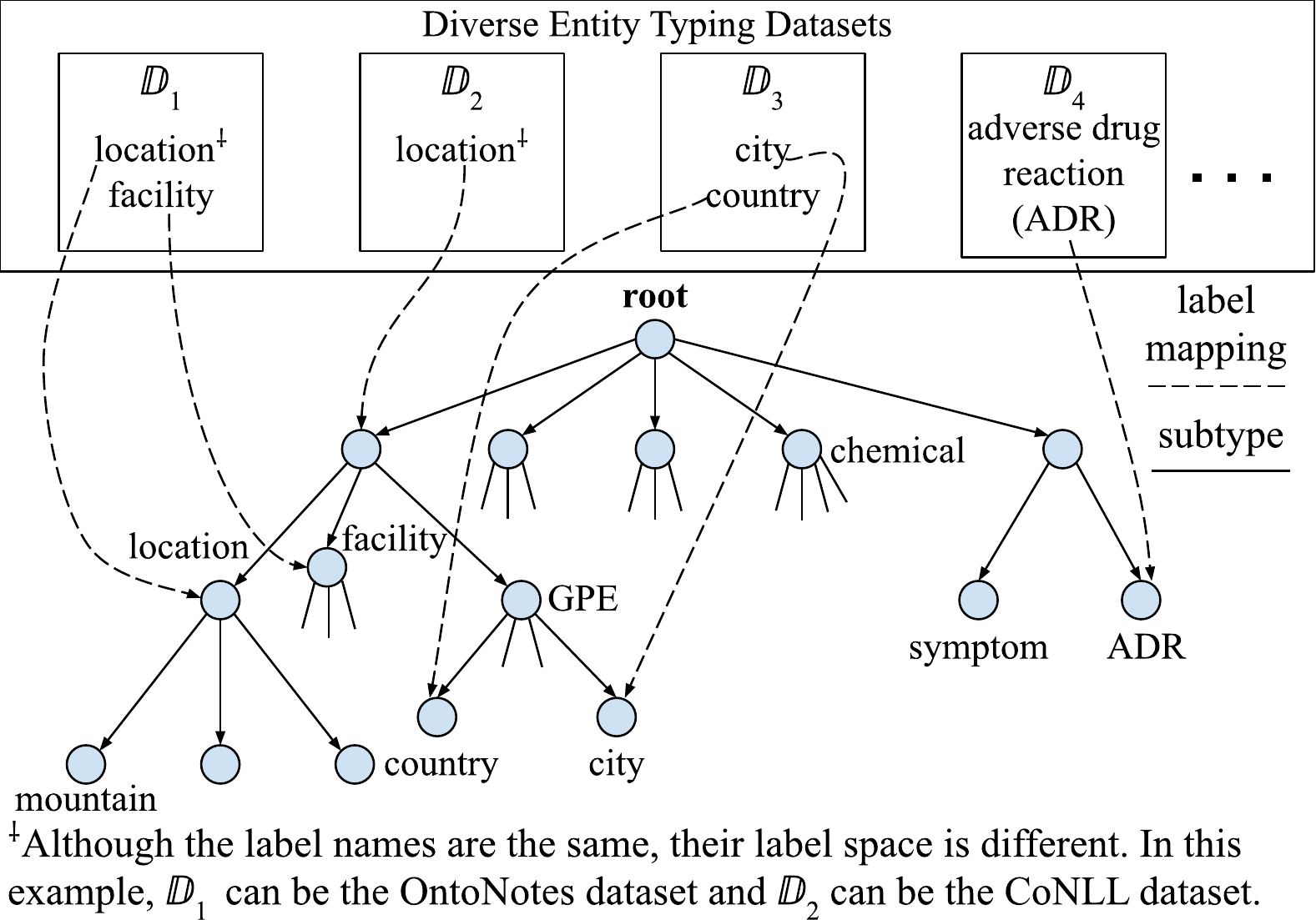}
\caption{A simplified illustration of the UHLS and the label mapping from individual datasets.}
\label{fig:unified_label_space}
\end{figure}

\subsection{Unified Hierarchy Label Set and Label Mapping}\label{sec:uhls}
The labels of entity mentions can be arranged in a hierarchy. For example, the label space of \textit{airports} is subsumed in the label space of \textit{facilities}. In literature, several hierarchies, such as WordNet \cite{miller1995wordnet} and ConceptNet \cite{liu2004conceptnet} exists. Even two ET datasets, BBN \cite{weischedel2005bbn} and Wiki organize labels in a hierarchy. However, none of these hierarchies can be directly used as discussed next. 

Our analysis of the labels of several ET datasets suggests that the presence of the same label name in the two or more datasets may not necessarily imply that their label spaces are same. For example, in the CoNLL dataset, the label space for the label \textit{location} includes facilities, whereas in the OntoNotes dataset \cite{weischedel2013ontonotes} the \textit{location} label space excludes facilities. These differences are because these datasets were created by different organizations, at different times and for a different objective. Figure \ref{fig:unified_label_space} illustrates this label space interaction.  Additionally, some of these labels are very specific to the domains, and not all of them are present in any publicly available hierarchies such as WordNet, ConceptNet or even knowledge bases (Freebase \cite{bollacker2008freebase} or WikiData \cite{vrandevcic2014wikidata}). 

Thus, to construct UHLS, we analyzed the annotation guidelines of several datasets and came up with an algorithm formally described in Algorithm \ref{algo:unified_hierarchy} and explained below. 

\begin{algorithm}[t]
 \DontPrintSemicolon
 \KwData{$\mathbb{Y} = \bigcup\limits_{i=1}^{n}\mathcal{Y}_{i}$}
 \KwResult{Unified Hierarchical Label Set (UHLS), $\mathcal{T}_{u} = (\mathcal{Y}_{u}, \mathcal{R}_{u}) $ and label mapping, $\phi$.}
 Initialize: $\mathcal{Y}_{u} = \{root\}, \mathcal{R}_{u} = \{\}$\;
 \For{$y \in \mathbb{Y}$}{
 	%\tcc{Case 2: }
    \eIf(\tcp*[f]{Case 2}){$\exists \mathcal{S} \subseteq \mathcal{Y}_{u}\; s.t.\; \mathcal{L}(y) == \mathcal{L}(\mathcal{S})$}{\label{algo:case2a}
    	$\phi(y) \mapsto \mathcal{S}$\;\label{algo:case2b}
    }
    (\tcp*[f]{Case 1}){
    	$v = \argminA\limits_{size(\mathcal{L}(v))} \{v \, | \, v \in \mathcal{Y}_{u} \, \& \, \mathcal{L}(y) \prec \mathcal{L}(v)\}$\;\label{algo:case1a}
        $\mathcal{Y}_{u} = \mathcal{Y}_{u} \cup \{y\}$\;%\tcc*[r]{Add y to UHLS}\;
        $\mathcal{R}_{u} = \mathcal{R}_{u} \cup \{(y, v)\}$\; \label{algo:case1b}
        $\phi(y) \mapsto y$\; 
        \For(\tcp*[f]{Update existing nodes}){$(x, v) \in \mathcal{R}_{u}$}{\label{algo:case1c}
        	\If{$x \neq y \, \& \, \mathcal{L}(x) \prec \mathcal{L}(y)$}{
                $\mathcal{R}_{u} = \mathcal{R}_{u} - \{(x, v)\}$\;
                $\mathcal{R}_{u} = \mathcal{R}_{u} \cup \{(x, y)\}$\;\label{algo:case1d}
            }
        }
        
        \For(\tcp*[f]{Restrict to tree hierarchy}){$\hat{v} \in \mathcal{Y}_{u}$}{\label{algo:case1e}
        	\If{$\mathcal{L}(\hat{v}) \prec \mathcal{L}(y) \, \& \, \hat{v} \notin subtree(y)$}{
        		$\phi(y) \mapsto \hat{v}$\;\label{algo:case1f}
        	}
        	
        }
    }
}

\caption{UHLS and label mapping creation algorithm.}
 \label{algo:unified_hierarchy}
\end{algorithm}

Given the set of all labels, $\mathbb{Y}$, the goal is to construct a type hierarchy, $\mathcal{T}_{u} = (\mathcal{Y}_{u}, \mathcal{R}_{u})$ and a label mapping $\phi : \mathbb{Y} \mapsto \mathcal{P}(\mathcal{Y}_{u})$. Here, $\mathcal{Y}_{u}$ is the set of labels present in the hierarchy, $\mathcal{R}_{u}$ is the relation set and $\mathcal{P}(\mathcal{Y}_{u})$ is the power set of the label set. To construct $\mathcal{T}_{u}$, we start with an initial type hierarchy, which can be $\mathcal{Y}_{u} = \{root\}, \mathcal{R}_{u} = \{\}$ or initialized by any existing hierarchy. We keep on processing each label $y \in \mathbb{Y}$ and decide if there is a need to update $\mathcal{T}_{u}$ and update the mapping $\phi$. For each label $y$ there are only two possible cases, either $\mathcal{T}_{u}$ is updated or not. \\
\textbf{Case 1, $\mathcal{T}_{u}$ is updated:} In this case $y$ is added to a child of an existing node in the $\mathcal{T}_{u}$, say $v$. While updating $\mathcal{T}_{u}$ it is ensured that $v = \argminA\limits_{size(\mathcal{L}(v))} \{v \; | \; v \in \mathcal{Y}_{u} \; \& \; \mathcal{L}(y) \prec \mathcal{L}(v)\;\} $, i.e., $\mathcal{L}(v)$ is the smallest possible label space that completely subsumes the label space of $y$ (lines \ref{algo:case1a}-\ref{algo:case1b}). After the update, if there are existing subtrees rooted at $v$, then if the label space of $y$ subsumes any of the subtree space, then $y$ becomes the root of those subtrees (lines \ref{algo:case1c}-\ref{algo:case1d}). In this case the label mapping is updated as $\phi(y) \mapsto y$, i.e., the label in an individual dataset is mapped to a same label name in UHLS. Additionally, if there exist any other nodes, $\hat{v} \in \mathcal{Y}_{u} \; s.t. \; \mathcal{L}(\hat{v}) \prec \mathcal{L}(y) \; \& \; \hat{v} \notin subtree(y)$, we add $\phi(y) \mapsto \hat{v}$ for all such nodes (lines \ref{algo:case1e}-\ref{algo:case1f}). This additional condition ensures that even in the cases where the actual hierarchy will be a directed acyclic graph, we restrict it to a tree hierarchy by adding additional mappings.\\
\textbf{Case 2, $\mathcal{T}_{u}$ is not updated:} In this case, $\exists \mathcal{S} \subseteq \mathcal{Y}\; s.t.\; \mathcal{L}(y) == \mathcal{L}(\mathcal{S})$, i.e, there exists a subset of nodes whose union of label space is equal to the label space of $y$. If $|\mathcal{S}| > 1$, intuitively this means that the label space of $y$ is a mixed space, and from some other datasets labels with finer label spaces were added to $\mathcal{Y}_{u}$. If $|\mathcal{S}| = 1$, this means that some other dataset added a label which has the same label space. In this case we will only update the label mapping as $\phi(y) \mapsto \mathcal{S}$ (lines \ref{algo:case2a}-\ref{algo:case2b}).

In Algorithm \ref{algo:unified_hierarchy} whenever a decision has to be made related to a comparison between two label spaces, we refer a domain expert. The expert makes the decision based on the annotation guidelines for the queried labels and using existing organization of the queried label space in WordNet or Freebase if the queried labels are present in these resources. We argue that since the overall size of $\mathbb{Y}$ is several order of magnitude less than the size of annotated instances ($\approx250 << \; \approx3 \times 10^{6}$), having a human in the loop preserves the overall semantic property of the tree, which will be exploited by a partial loss function to enable fine-grained prediction across domains. An illustration of UHLS and label mapping is provided in Figure \ref{fig:unified_label_space}.

In the next section, we will describe how the UHLS and the label mapping will be used by a learning model to make finest possible predictions across datasets.

\subsection{Learning Model}
Our learning model can be decomposed into two parts: (1) Neural Mention and Context Encoders to encode the entity mention and its surrounding context into a feature vector; (2) Unified Type Predictor to infer entity types in the UHLS.
\subsubsection{Neural Mention and Context Encoder}\label{sec:context_encoder}
The input to our model is a sentence with the start and end index of entity mentions. Following the work of \cite{shimaoka2017neural,abhishek2017fine,xu2018neural} we use Bi-directional LSTMs \cite{graves2013speech} to encode left and right context surrounding the entity mention and use a character level LSTM to encode the entity mention. After this we concatenate the output of the three encoders, to generate a single representation ($R$) for the input.

\subsubsection{Unified Type Predictor}\label{sec:utp}
Given the input representation, $R$, the objective of the predictor is to assign a type from the unified label set $\mathcal{Y}_{u}$. % However, the input datasets label are from the set $\mathbb{Y}$.
Thus, during model training, using the mapping function $\phi: \mathbb{Y} \mapsto \mathcal{P}(\mathcal{Y}_{u})$ we convert individual dataset specific labels to the unified label set, $\mathcal{Y}_{u}$. Due to one to many mapping, now there are multiple positive labels available for each individual input label $y$. Lets call the mapped label set for an input label $y$ as $\mathcal{Y}_{m}$. Now, if any of the mapped label $\hat{y} \in \mathcal{Y}_{m}$ has descendants, then the descendants are also added to $\mathcal{Y}_{m}$\footnote{This is exempted when the annotated label is a coarse label and a fine label from the same dataset exist in the subtree.}. For example, if the label \textit{GPE} from the OntoNotes dataset, is mapped to the label \textit{GPE} in the UHLS, then \textit{GPE} as well as all descendants of \textit{GPE} are possible candidates. This is because, even though the original example in OntoNotes is a name of a city, the annotation guidelines restrict the fine-labeling. Thus the mapped set would be updated to \{\textit{GPE}, \textit{City}, \textit{Country}, \textit{County}, ...\}. Additionally, some label have a one-to-many mapping, for example, for the label \textit{MISC} in CoNLL dataset, the candidate labels could be \{\textit{product}, \textit{event}, ...\}. 

From the set of mapped candidate labels, a partial label loss function will select the best candidate label. Due to the inherent design of the UHLS and label mapping, there will always be examples available that will be mapped only at a single leaf node. Thus allowing fine labels in the candidate set for actual coarse labels, will encourage model to predict finer labels across datasets.

\subsubsection{Partial Hierarchical Label Loss}\label{sec:partial_loss}
A partial label loss deals with the situation where training example have a set of candidate labels and among which only a subset is correct for that given example \cite{nguyen2008classification,cour2011learning,zhang2017disambiguation}. 

In our case, this situation arises because of the mapping of the individual dataset labels to the UHLS. We use a hierarchy aware partial loss function as proposed in \cite{xu2018neural}. We first compute the probability distribution for the labels available in $\mathcal{Y}_{u}$ as described in equation 1. Here $W$ is a weight matrix of size $|R| \times |\mathcal{Y}_{u}|$ and $x$ is the input entity mention along with its context.
\begin{equation}
p(y|x) = softmax(RW + b)
\end{equation}
Then we compute $\hat{p}(y|x)$, a distribution adjusted to include a weighted sum of the ancestors probability for each label as defined in equation 2. Here $\mathcal{A}_{t}$ is the set of ancestors of the label $y$ in $\mathcal{R}_{u}$ and $\beta$ is a hyperparameter. 
\begin{equation}
\hat{p}(y|x) = p(y|x) + \beta * \sum_{t \in \mathcal{A}_{t}} p(t|x)
\end{equation}
Then we normalize $\hat{p}(y|x)$. From this normalized distribution, we select a label which has the highest probability and is also a member of the mapped labels $\mathcal{Y}_{m}$. We assumed the selected label to be correct and propagate the log-likelihood loss. The intuition behind this is that given the design of the ULHS and label mapping; there will always be examples where $\mathcal{Y}_{m}$ will contain only one element, in that case, the model gets trained for that label. In the case where there are multiple labels, the model has already built a belief about the fine label suitable for that example because of simultaneously training with inputs having a single mapped label. Restricting that belief to the mapped labels encourages correct fine-predictions for these coarsely labeled examples.

\setlength{\belowcaptionskip}{1pt}

\begin{table}[t]
\centering
\resizebox{\columnwidth}{!}{%
\begin{tabular}{p{1.7cm}p{3.2cm}p{1.0cm}p{1.25cm}p{0.9cm}}
\toprule
\textbf{Dataset}	                    & \textbf{Domain}        					    & \textbf{No. of Labels}& \textbf{Mention count}& \textbf{Fine labels} \\ \toprule
BC5CDR \cite{li2016biocreative}         & Clinical abstracts			 	            & 2      				& 9,385        			& No    \\ 
CoNLL \cite{tjong2003introduction}   	& Reuters news stories                     		& 4      				& 23,499     			& No    \\
JNLPBA \cite{kim2004introduction}   	& Life sciences abstracts 	 	                & 5      				& 46,750       			& Yes   \\
CADEC \cite{karimi2015cadec}    		& Medical forum              					& 5      				& 5,807       	 		& Yes   \\
OntoNotes \cite{weischedel2013ontonotes}& Newswire, conversations, newsgroups, weblogs 	& 18     				& 1,16,465     			& No    \\
BBN \cite{weischedel2005bbn}      		& Wall Street Journal text                      & 73     				& 86,921      			& Yes   \\
Wiki \cite{ling2012fine}      			& Wikipedia			       					 	& 116    				& 20,00,000        		& Yes   \\ \hline
\end{tabular}}
\caption{Description of the seven ET datasets.}
\label{tab:dataset_stats}
\end{table}

\setlength{\belowcaptionskip}{-7pt}

\begin{figure*}[t]
\centering
\includesvg[width=0.8\textwidth]{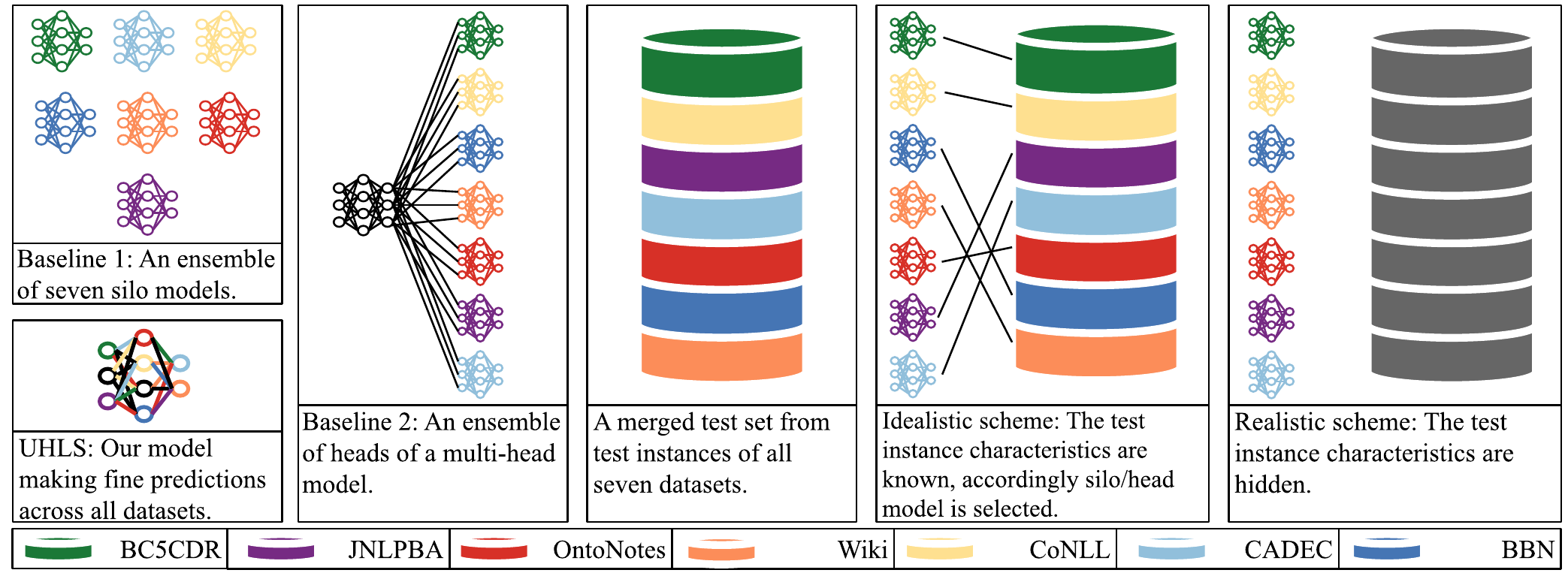}
\caption{A pictorial illustration of the complete experimental setup.}
\label{fig:complete_experiment_setup}
\end{figure*}

\section{Experiments and Analysis}
In this section, we describe the datasets used, details of experiments related to UHLS creation, baseline models, model training, evaluation schemes and result analysis. 
% The source code to reproduce the results will be available at \url{https://github.com/_____/_____}.
\subsection{Datasets}
Table \ref{tab:dataset_stats} describes the seven datasets used in this work. These datasets are diverse, as they span several domains, none of them have an identical label set and some datasets capture fine-grained labels while others only have coarse labels. Also, the Wiki \cite{ling2012fine} dataset is automatically generated using distant supervision process \cite{craven1999constructing} and has multiple labels per entity mention in its label set. The other remaining datasets have a single label per entity mention.

\subsection{UHLS and Label Mapping}
We followed the Algorithm \ref{algo:unified_hierarchy} to create the UHLS and the label mapping. To reduce the load on domain experts for verification of the label spaces, we initialized the UHLS with the BBN dataset hierarchy. We keep on updating the initial hierarchy until all the labels from the seven datasets were processed. There were total $223$ labels in $\mathbb{Y}$ and in the end $\mathcal{Y}_{u}$ had $168$ labels. This difference in label count is due to the mapping of several labels to one or multiple existing nodes, without the creation of a new node. This corresponds to case 2 of the UHLS creation process (lines \ref{algo:case2a}-\ref{algo:case2b}, Algorithm \ref{algo:unified_hierarchy}). Also, this indicates the overlapping nature of the seven datasets. The label set overlap is illustrated in Figure \ref{fig:label_overlap}. The \textit{MISC} label from CoNLL dataset has the highest ten number of mappings to the UHLS nodes. Wiki and BBN datasets were the largest contributor towards fine labels with $96$ and $57$ labels at the leaf of UHLS. However, only $25$ fine-grained labels were shared by these two datasets. This indicates that even though these are the fine-grained datasets with one of the largest label sets, each of them has complementary labels.

\subsection{Baselines}
%We compared our learning model performance with two baseline models under several evaluation schemes described later. The first baseline is a learning model trained only on a single dataset. We name these as silo models. In this baseline, the input is fed through a mention and context encoder, and the output labels are the same as that was available in the original dataset. In the case of a single label dataset, we use a standard softmax based cross-entropy loss. For multi-label datasets, we use a sigmoid based cross-entropy loss.

We compared our learning model with two baseline models. The first baseline is an ensemble of seven learning models, where each model is trained on one of the seven datasets. We name this model a silo ensemble model\footnote{Here unlike traditional ensemble models, in silo ensemble, the learning models are trained on different datasets.}. In this ensemble model, each silo model has the same mention and context encoder structure described in Section \ref{sec:context_encoder}. However, the loss function is different. For single-label datasets, we use a standard softmax based cross-entropy loss. For multi-label datasets, we use a sigmoid based cross-entropy loss.

The second baseline is a learning model trained using a classic hard parameter sharing multi-task learning framework \cite{caruana1997multitask}. In this baseline, all the seven datasets are fed through a common mention and context encoder.  For each dataset, there is a separate classifier head with the output labels same as that was available in the respective original dataset. We name this baseline as a multi-head ensemble baseline\footnote{Here since the ``task" is the same, i.e., entity typing, we use the term multi-head instead of multi-task for the baseline.}. Similar to the silo models, the appropriate loss function is selected for each head. The only difference between the silo and multi-head model is the way mention and context representations are learned. In the multi-head model, the representations are shared across datasets. In silo models, the representations are learned separately for each dataset.

%Both of these baselines use the same mention and context encoder architecture as used by our model, i.e., the LSTM based encoders. 

\subsection{Model Training}
For each of the seven datasets, we use the standard train, validation and testing split. If the standard splits are not available, we randomly split the available data into $70$\%, $15$\%, and $15$\%, and use them as train, validation, and testing set respectively. In the case of the silo model, for each dataset, we train a model on its training split and select the best model using its validation split. In the case of the multi-head and our proposed model, we train the model on the training splits of all seven datasets together and select the best model using the combined validation split.\footnote{The source code and the implementation details are available at: \url{https://github.com/abhipec/ET_in_the_wild}}.

\subsection{Experimental Setup}\label{sec:evaluation_setting}
%In our experimental setup, we have seven silo models, a multi-head model, and our proposed model. We compare these models under two evaluation measures, an \textbf{idealistic} and a \textbf{realistic} measure. In each of these measure, we use two evaluation metrics, a best effort metric and a fine-grained prediction metric.
Figure \ref{fig:complete_experiment_setup} illustrates the complete experimental setup along with the learning models compared. In this setup, the objective is to measure the learning model's generalizability for the ET task as a whole, rather than on any specific dataset. To achieve this, we merged the test instances from the seven datasets listed in Table \ref{tab:dataset_stats} to form a combined test corpus. On this test set, we compared the performance of the baseline models with the learning model trained via our proposed framework. We compare these models performance using the following evaluation schemes. 
\subsubsection{Evaluation schemes}
%An \textbf{idealistic measure} assumes that a single dataset is a representative of the ET world. The existing work in ET only evaluates learning models in this narrowly focused measure. We name this measure idealistic because we know during the testing information about the test data domain and the candidate label set.  In this measure, given a test dataset, we pick a silo model (or head of the multi-head model) which has been trained on a training dataset with the same characteristics as the test dataset. In this measure, for the silo and multi-head models, if the source label set is coarse-grained, then the predictions are always limited to the coarse label set. In our proposed model, there is no selection step, i.e., all seven test datasets are indistinguishable, as there is only one classifier which always predicts in the UHLS across all datasets. 
\noindent
\textbf{Idealistic scheme:} Given a test instance, this scheme picks a silo model from the silo ensemble model (or head of the multi-head ensemble model) which has been trained on a training dataset with the same domain and target labels set as the test instance. This scheme gives an advantage to the ensemble baselines and compares the models in the traditional ways.

\noindent
\textbf{Realistic scheme:} In this scheme,  all of the test instances are indistinguishable in their domain and candidate label set. In other words,  given a test instance, learning models do not have information about its domain and target labels. This is a challenging evaluation scheme and close to real-world setting, where once learning models are deployed, it cannot be guaranteed that the user submitted test instances will be from the same domain. In this scheme, the silo ensemble and multi-head ensemble models assign a label to a test instance based on the following criteria:

%A \textbf{realistic measure} assumes that a single dataset is a representative of only a small subset of the ET world. We name this measure realistic because we don't have any information about the test data domain and the candidate label set\footnote{Here our assumption is that the collection of all of the domains and the label sets present in the available datasets is a representative of the ET world. A test data can have any combination of domain/label set from this world. However, we don't know about the exact domain/label set.}. For the silo and multi-head models, the missing information creates a major issue of type assignment.  We have multiple silo/head models and for a test entity mention, we need to assign a type from one of these models. In this measure, we pass every test example through all of the silo models and all of the heads of the multi-head model. Each model/head will produce a ranked list of its label with a confidence score. We pick a final label based on the two schemes described below:

\noindent
\textbf{Highest confidence label (HCL):}  The label which has the highest confidence score among the different models/heads of an ensemble model. For example, let there be two models/heads, MA and MB, in a silo/multi-head ensemble model. For a test instance, MA assigns the score of 0.1, 0.2 and 0.7 for the labels $l_{1}$, $l_{2}$ and $l_{3}$ respectively. For the same test instance, MB assigns the score of 0.05 and 0.95 for the labels $l_{4}$ and $l_{5}$ respectively. Then the final label will be the label $l_{5}$ which has a confidence score of 0.95.

\noindent
\textbf{Relative highest confidence label (RHCL):} The label which has the highest normalized confidence score among the different models/heads from an ensemble model. Continuing with the example mentioned above for MA and MB, in this criteria, we normalize the confidence score for each model based on the number of labels the model is predicting. In this example, MA is predicting three labels and MB is predicting two labels. Here the normalized scores for MA will be 0.3, 0.6 and 2.1 for the label $l_{1}$, $l_{2}$, and $l_{3}$ respectively. Similarly, the normalized scores for MB will be 0.1 and 1.9 for the label $l_{4}$ and $l_{5}$. Then the final label will be the label $l_{3}$ with the confidence score of 2.1. 
%\textbf{Highest confidence label (HCL):} We pick a label which has the highest confidence score across all model/head for a test example. If there are ties, then these are resolved using the RHCL scheme.\\
%\textbf{Relative highest confidence label (RHCL):} In this scheme, we re-rank the model/head confidence relative to the random chance. For example, if the model/head is making a prediction from a label set of three with confidence [0.1, 0.2, 0.7], we adjust the confidence of each label prediction relative to the 33.33\% chance. The new scores will be [0.3, 0.6, 2.1]. After re-ranking, we pick a label which has the highest relative confidence score across all models/heads for a test example.

%Recall that the experimental setup includes multiple models, each having a different label set. The existing classifier integration strategies \cite{zhou2012ensemble}, such as sum rule or majority voting are not suitable for this work because every classifier has a different label set. 

Recall that the experimental setup includes multiple models, each having a different label set. The existing classifier integration strategies \cite{zhou2012ensemble}, such as sum rule or majority voting are not suitable in this setup. For these evaluation schemes, we use the evaluation metrics described in the following section. 

\subsection{Evaluation metrics}\label{sec:evaluation_metric}
%In the evaluation measure, there are cases where the label set of the model's prediction does not match with the label set of the gold dataset. Due to this reason, without re-annotating test portion of all the datasets, we cannot have an exhaustive comparison among models. To overcome this issues, we propose two evaluation metrics through which a comparison can be made with minimum re-annotation effort.

In the evaluation schemes, there are cases where the predicted label is not part of the gold dataset label set. For example, our proposed model or the ensemble model might predict a label \textit{city} for a test instance which has a gold label annotated as a \textit{geopolitical entity}. Here, the models are predicting a fine-grained label, however, the dataset from where the test instance came only had annotations at the coarse level. Thus, without manually verifying, it is not possible to know whether the model's prediction was correct or not. To overcome this issue, we propose two evaluation metrics, which allows us to compare learning models making predictions in different label sets with minimum re-annotation effort.

\begin{figure}[t]
\centering
\includesvg[width=1.0\columnwidth]{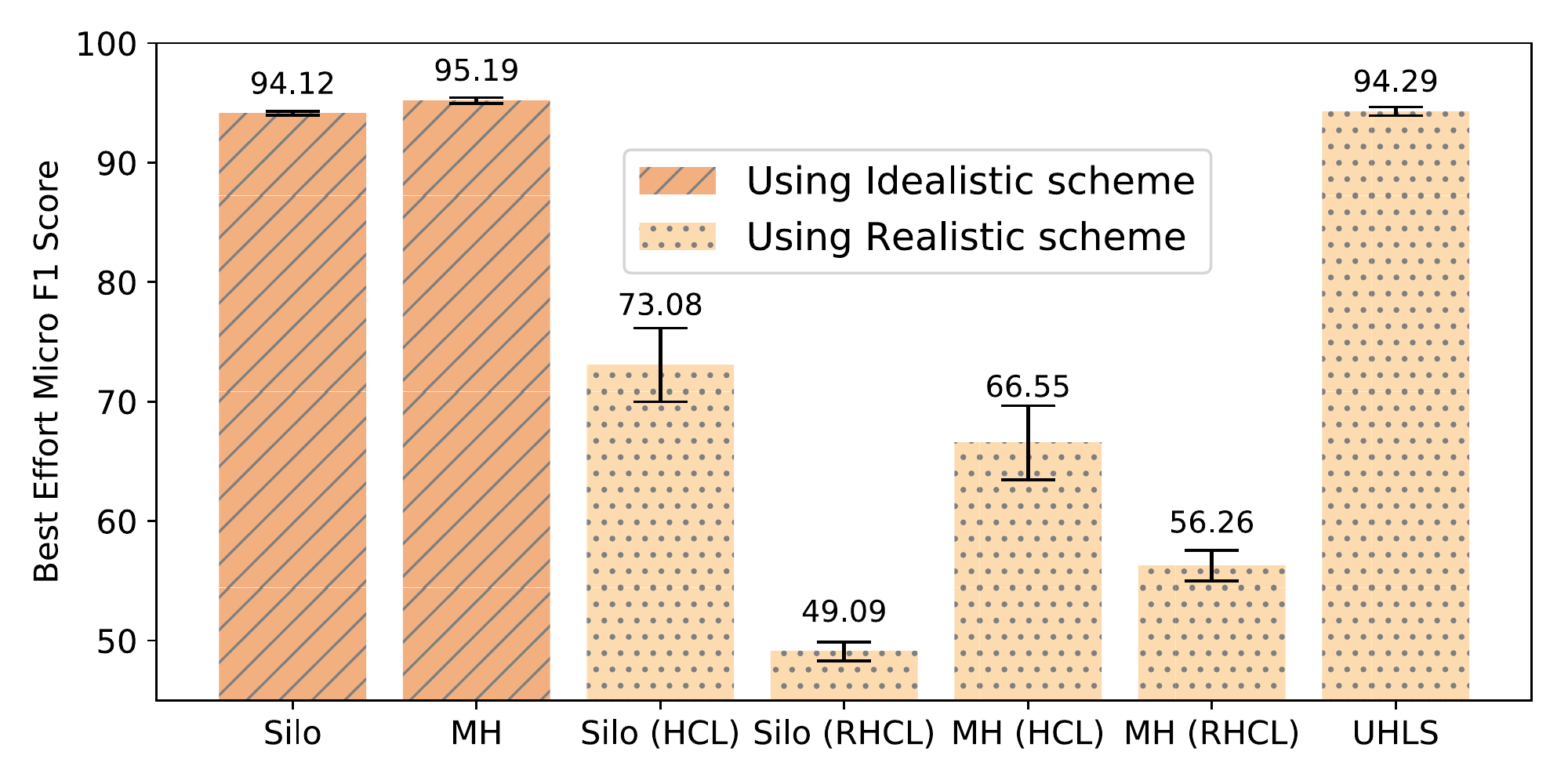}
\caption{Comparison of learning models in the idealistic and realistic schemes.}
\label{fig:idelistic_realistic}
\end{figure}

In the first metric, we compute an aggregate micro-averaged F1 score on best effort basis. It is based on the intuition that if the labels are only annotated at a coarse level in the gold test annotations, then even if a model predicts a fine-label within that coarse label, this metric should not penalize such cases\footnote{Exception is where the source dataset also has fine-grained labels.}. To find the fine-coarse subtype information, we use the UHLS and the label mapping. We map both prediction and gold label to the UHLS and evaluate in that space. We compute this metric both in an idealistic and realistic scheme. By design, this metric will not capture errors made at a finer level, which the next metric will capture.

%In the second metric, we measure how good are the fine-grained predictions on examples where the gold dataset has only coarse labels. For example, if the model's prediction was \textit{city} on a dataset where all cities and countries are clubbed together under a label \textit{GPE}. We re-annotate a representative sample of a coarse-grained dataset and evaluate the model's performance on this sample.
In the second metric, we measure how good are the fine-grained predictions on examples where the gold dataset has only coarse labels. We re-annotate a representative sample of a coarse-grained dataset and evaluate the model's performance on this sample.
\subsection{Result and Analysis}
\subsubsection{Analysis of the idealistic scheme results}
%In this scheme, we can observe that when the information about the test data characteristics is known, the multi-head model outperforms our proposed approach and the silo models. The primary reason could be that the model has learned better shared representations using the multi-task framework as well as has a independent head for each dataset to learn dataset specific idiosyncrasy. In our proposed model, the overall complexity including the label search space increases compared to both multi-head and silo models. Despite of this increased complexity, we can observe that its performance is competitive. 

In Figure \ref{fig:idelistic_realistic}, we can observe that the multi-head ensemble model outperforms the silo ensemble model ($95.19\%$ vs. $94.12\%$). The primary reason could be that the multi-head model has learned better representations using the multi-task framework as well as has an independent head for each dataset to learn dataset specific idiosyncrasy. The performance of our single model (UHLS) is between the silo ensemble model and multi-head ensemble model. Note that this performance comparison is in a setting which is the best possible case for ensemble models where the ensemble models know complete information about the test instance domain and label set. Despite this, UHLS model which does not require any information about test instance domain and candidate labels performs competitive ($94.29\%$), even better than the silo ensemble model. Moreover, the ensemble models do not always predict the finest possible label, whereas UHLS can ($\S$ \ref{sec:fine_grained_results}).
\subsubsection{Analysis of the realistic scheme results}
In Figure \ref{fig:idelistic_realistic}, we can observe that both silo ensemble and multi-head ensemble model performs poorly in this scheme. The best result for ensemble models ($73.08\%$) is obtained by the silo ensemble model when the labels were assigned using the HCL criteria. We analyzed some of the outputs of ensemble models and found that there were several cases where a narrowly focused model predicts with very high confidence (0.99 probability or above) out-of-scope labels. For example, prediction of label ADR with confidence 0.999 by a silo model trained on the CADEC dataset for a \textit{sports event} test instance of Wiki domain. The performance of our UHLS model is $94.29\%$, which is an absolute improvement of $21.21\%$ compared to the next best model Silo (HCL) model in the realistic scheme of evaluation. 

\subsubsection{Analysis of the fine-grained predictions}\label{sec:fine_grained_results} 
\begin{figure}[t]
\centering
\includesvg[width=1.0\columnwidth]{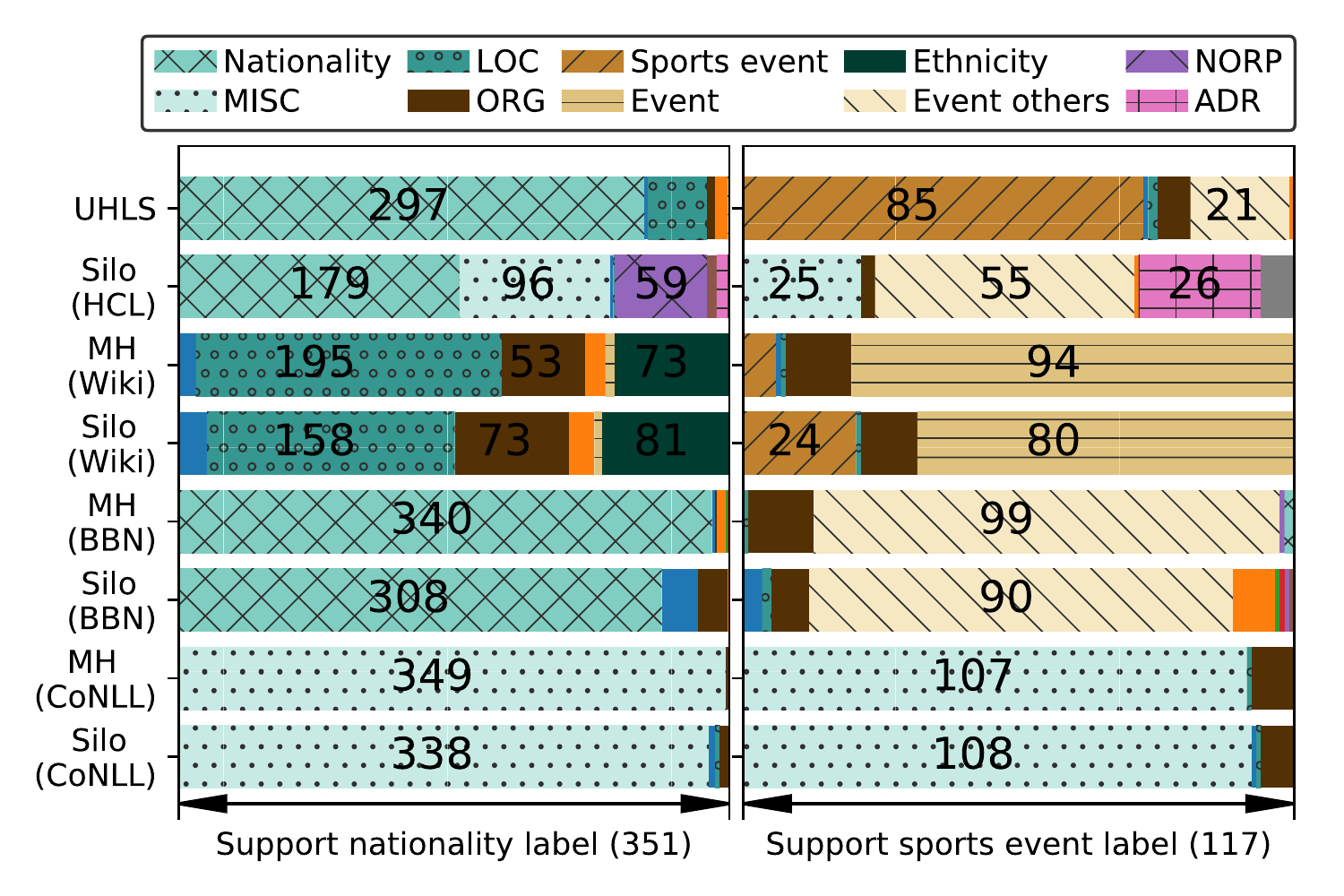}
\caption{Analysis of Fine-grained label predictions. The two columns specify results for nationality and sports event label. Each row represents a model used for prediction. The results can be interpreted as, out of 351 entity mentions with type nationality, model Silo (CoNLL) predicted 338 as MISC type and the remaining as other types illustrated.}
\label{fig:fine_result_analysis}
\end{figure}

For this analysis, we re-annotate the examples of type \textit{MISC} from the CoNLL test set into \textit{nationality} (support of 351), \textit{sports event} (support of 117) and others (support 234). We analyzed the prediction of different models for the \textit{nationality} and \textit{sports event} labels. Note that this is an interesting evaluation where the test instances domain is Reuters News, and the in-domain dataset does not have labels \textit{nationality} and \textit{sports event}. The \textit{nationality} label is contributed by the BBN dataset whose domain is Wall Street Journal. The \textit{sports event} label is contributed by the Wiki dataset whose domain is Wikipedia. The results (Figure \ref{fig:fine_result_analysis}) are categorized into three parts as described below:

\noindent
\textbf{In-domain results:} The bottom two rows, Silo (CoNLL) and MH (CoNLL) represent these results. We can observe that in this case, since train and test dataset are from the same domain, these models can predict accurately the label \textit{MISC} for both the \textit{nationality} and \textit{sports event} instances. However, \textit{MISC} is not a fine-grained label. These results are from the idealistic scheme where it is known about the test instance characteristics. 

\noindent
\textbf{Out of domain but with known candidate label:} The middle four rows, Silo (BBN), MH (BBN), Silo (Wiki) and MH (Wiki) represent these results. In this case, we assume that the candidate labels are known, and pick the models which can predict that label. However, there is not a single silo/head model in the ensemble models which can predict both \textit{nationality} and \textit{sports event} labels. For example, model/head with the BBN label set can predict the label \textit{nationality} but not the \textit{sports event} label. For \textit{sports event} instances, it assigns a coarse label \textit{events other}, which also subsumes other events such as \textit{elections}. Similarly, model/head with the Wiki label set can predict the label \textit{sports event} but not the label \textit{nationality}. For \textit{nationality} instances, it assigns completely out of scope labels such as \textit{location} and \textit{organizations}. The out of scope predictions are due to the domain mismatch. 

\noindent
\textbf{No information about domain or candidate label:} The top two rows, Silo (HCL) and UHLS represent these results. The Silo (HCL) is a silo ensemble model with the realistic evaluation scheme. We can observe that this model makes out of scope predictions such as predicting \textit{ADR} for \textit{sports event} instances. The UHLS model is trained using our proposed framework. It can predict fine-grained labels in both \textit{nationality} and \textit{sports event} test instances, even though two different datasets contributed these labels. Also, it does not use any information about the test instance domain or candidate labels.

\subsubsection{Example output on different datasets}

\begin{figure}[t]
\centering
\includesvg[width=\columnwidth]{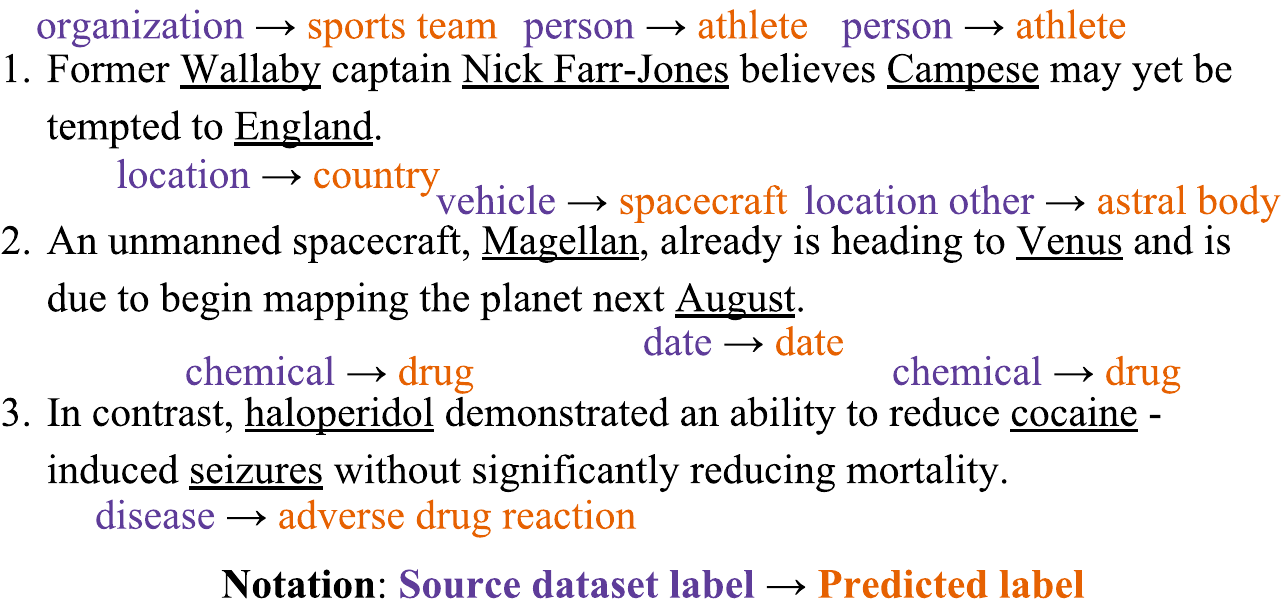}
\caption{Example output of our proposed approach. Sentence 1, 2, 3 are from the CoNLL, BBN and BC5CDR dataset respectively.}
\label{fig:example_output}
\end{figure}

In Figure \ref{fig:example_output}, we show the labels assigned by the model trained using the proposed framework on the sentences from the CoNLL, BBN and BC5CDR datasets. We can observe that, even though the BBN dataset is fine-grained, it has complementary labels compared with the Wiki dataset. For example, for the entity mention \texttt{Magellan}, a label \textit{spacecraft} is assigned. \textit{Spacecraft} label is only present in the Wiki dataset. Additionally, even in sentences from clinical abstracts, the proposed approach is assigning fine-types, which came from a dataset with the medical forum domain. For example, \textit{ADR} label is only present in the CADEC dataset with the domain of medical forum. The proposed approach can aggregate fine-labels across datasets and makes unified fine-grained predictions.

\subsubsection{Result and analysis summary}
Collective learning framework allows a limitation of one dataset being covered by some other dataset(s). Our results convey that a model trained using CLF on an amalgam of diverse datasets generalizes better for the ET task as a whole. Thus, the framework is suitable for the ET in the wild problem.

\section{Related Work}
To the best of our knowledge, the work of \cite{redmon2017yolo9000} in the visual object recognition task is closet to our work. They consider two datasets. First a coarse-grained and second, a fine-grained. Label set of the first dataset is assumed to be subsumed by the label set of the second dataset. Thus coarse-grained labels can be mapped to fine-grained dataset labels in a one-to-one mapping. Additionally, they did not propagate the coarse labels to the finer labels. As demonstrated by our experiments, when several real-world datasets are merged, one to one mapping is not possible. In our work, we provide a principled approach where multiple datasets can contribute to fine-grained labels. In our framework, a partial loss function enables fine-label propagation on datasets with coarse labels. 

In the area of cross-lingual syntactic parsing, there is a notation of universal POS tagset \cite{petrov2012universal}. This tagset is a collection of coarse tags that exist in similar form across languages. Utilizing this tagset and a mapping from language-specific fine-tags, it becomes possible to train a single model in a cross-lingual setting. In this case, the mapping is many-to-one, i.e., a fine-category to a coarse category, thus the models are limited to predict a coarse-grained label.

Related to the use of partial label loss function in the context of the ET problem, there exist other notable works including \cite{ren2016afet} and \cite{abhishek2017fine}. In our work, we use the current state-of-the-art hierarchical partial loss function proposed in \cite{xu2018neural}.

\section{Conclusion}
%The key idea of our paper is that by using in conjunction, a UHLS, one-to-many label mappings, and a partial loss function; we can train a single classifier on several diverse datasets together. The single classifier generalize better for the ET tasks as a whole, and predicts finest possible label across all the diverse ET datasets.

%Our analysis indicates the following observations. First, focusing on each dataset as a world of ET is not suitable for the real-world purpose. Second, in a real world, any combination of domain and label set is possible. Third, cases where source label is a very coarse-grained label, a one-to-many mapping helps to assign finer-labels. Fourth, even if there are two large fine-grained datasets, BBN and Wiki, they both provide complementary information. Fifth, our proposed models enables fine-grained predictions across all datasets by capitalizing on the knowledge learned from labels coming from diverse datasets.  

In this paper, we propose building learning models that generalize better on the ET as a whole, rather than on a specific dataset.  We comprehensively studied ET in the wild task which includes problem definition, collective learning framework, and evaluation setup. We demonstrated that by using in conjunction a UHLS, one-to-many label mappings, and a partial hierarchical loss function; we can train a single classifier on several diverse datasets together. The single classifier collectively learns from diverse datasets and predicts the best possible fine-grained label across all datasets, outperforming an ensemble of narrowly focused models in their best possible case. Also, during collective learning there is a multi-directional knowledge flow, i.e., there is no one source or target dataset. This knowledge flow is different from the well studied multi-task and transfer learning approaches \cite{pan2010survey} where the prime objective is to transfer knowledge from a source dataset to a target dataset.

In NLP there are several tasks such as entity linking \cite{shen2014entity}, relation classification \cite{hendrickx2009semeval}, and named entity recognition \cite{nadeau2007survey}, where the current focus in on excelling at a particular dataset, not on a particular task. We expect that collective learning approaches will open up a new research direction for each of these tasks.
%Some of these tasks such as relation classification have similar characteristics to that of the ET task, where the objective is to assign a label to a given input. For these tasks, our proposed CLF can be directly used, whereas other tasks may require suitable modifications. 

\bibliographystyle{ACM-Reference-Format}
\bibliography{sample-base}

\end{document}